\documentclass[a4paper,fleqn]{cas-dc}

\usepackage[numbers]{natbib}
\usepackage{float}
\usepackage{graphicx}
\usepackage{caption}
\usepackage{times}
\usepackage{xcolor}
\usepackage{color}
\usepackage[switch]{lineno}

\usepackage{algorithm,algpseudocode,algorithmicx}
\usepackage{subfigure}
\usepackage{booktabs}
\usepackage{pgfgantt}
\usepackage{amsfonts}
\usepackage{amsmath}
\usepackage[justification=centering]{caption}
\usepackage{makeidx}
\makeindex

\begin{document}
\let\WriteBookmarks\relax
\def\floatpagepagefraction{1}
\def\textpagefraction{.001}
\shorttitle{Efficient Search of Comprehensively Robust Neural Architectures via Multi-fidelity Evaluation}
\shortauthors{Jialiang Sun et~al.}

\title [mode = title]{Efficient Search of Comprehensively Robust Neural Architectures via Multi-fidelity Evaluation}                    
\author[1]{\color{black}Jialiang Sun}

\author[1]{\color{black}Wen Yao}
\cormark[1]
\author[1]{\color{black}Tingsong Jiang}
\cormark[1]

\author[1]{\color{black}Xiaoqian Chen}

%
%
\address[1]{Defense Innovation Institute, Chinese Academy of Military Science, Beijing 100071, China}

\cortext[cor1]{Corresponding author}

\begin{abstract}
Neural architecture search (NAS) has emerged as one successful technique to find robust deep neural network (DNN) architectures. However, most existing robustness evaluations in NAS only consider $l_{\infty}$ norm-based adversarial noises. In order to improve the robustness of DNN models against multiple types of noises, it is necessary to consider a comprehensive evaluation in NAS for robust architectures. But with the increasing number of types of robustness evaluations, it also becomes more time-consuming to find comprehensively robust architectures. To alleviate this problem, we propose a novel efficient search of comprehensively robust neural architectures via multi-fidelity evaluation (ES-CRNA-ME). Specifically, we first search for comprehensively robust architectures under multiple types of evaluations using the weight-sharing-based NAS method, including different $l_{p}$ norm attacks, semantic adversarial attacks, and composite adversarial attacks. In addition, we reduce the number of robustness evaluations by the correlation analysis, which can incorporate similar evaluations and decrease the evaluation cost. Finally, we propose a multi-fidelity online surrogate during optimization to further decrease the search cost. On the basis of the surrogate constructed by low-fidelity data, the online high-fidelity data is utilized to finetune the surrogate. Experiments on CIFAR10 and CIFAR100 datasets show the effectiveness of our proposed method.
\end{abstract}




\maketitle
\section{Introduction}
Deep neural network (DNN) has achieved notable success in various computer vision tasks, including image classification \cite{lu2007survey,wang2017residual}, object detection \cite{szegedy2013deep,zhao2019object}, and semantic segmentation \cite{QingAdaptive2022}. However, the majority of works have proved that adding the specific perturbations can make DNN output wrong results \cite{pgd2018,li2022approximated,li2023adaptive}. The images with perturbations are called adversarial examples (AEs), which can bring a huge threat to the real-world application of DNN, such as auto driving and face recognition systems. Therefore, improving the robustness of DNN has attracted the increasing attention of researchers.

Over the past years, to improve the robustness of DNN against AEs, various defensed methods are developed, such as the defensive distillation \cite{papernot2016distillation}, dimensionality reduction \cite{chattopadhyay2022robustness}, input transformations \cite{bhagoji2018enhancing}, and adversarial training \cite{tramer2017ensemble,shafahi2019adversarial,andriushchenko2020understanding}. Recently, intrinsically robust architectures have been increasing researchers' attention. Designing intrinsically robust architectures can also easily combine other defense techniques, such as adversarial training to further enhance the robustness of DNN models. However, manually-design needs to take a large burden, and the designed architecture may not possess superior performance. As an emerging tool, neural architecture search (NAS) \cite{elsken2019neural,yao2020efficient,pham2018efficient} have been developed to help automatically find more robust architectures. NAS can not only release human labor but also improve the performance of models.

The core objective of NAS is to find near-optimal neural architectures by searching in a predefined model search space. During the search process, the key procedure includes defining the search space, performance evaluation, and conducting the search strategy. In the NAS works, the adopted strategies are evolutionary algorithm \cite{liu2021survey}, bayesian optimization \cite{white2021bananas}, reinforcement learning \cite{zoph2016neural}, gradient-based optimization \cite{liu2018darts}, and weight sharing based search algorithm \cite{chu2021fairnas}. The gradient-based optimization method is differentiable, while others are non-differentiable. Among these NAS methods, the process of the weight-sharing-based search algorithm includes training the supernet consisting of all possible sub-network and searching in the trained supernet, which can reduce the computational burden of training each candidate architecture from scratch.

With the rapid development of NAS techniques, more recent works have been devoted to improving the robustness of DNN models. The earliest work about NAS for robust architectures is RobNet, proposed by Guo et al. \cite{whencvpr}, which can find more robust architectures under projected gradient descent (PGD) attack using weight sharing based architecture search method. To improve the performance of the searched architectures, many search strategies are also combined into NAS for robust architectures, which can be categorized into differentiable and non-differentiable based. Among the former methods, Hosseini et al. \cite{Differentiable2021} utilized the differentiable architecture search method to efficiently find robust architectures by taking the $F$ norm of the Jacobian matrix as the robustness evaluation. Mok et al. \cite{AdvRush2021} further considered the $F$ norm of the Hessian matrix in the evaluation of differentiable NAS. Among the latter methods, Liu et al. \cite{LiuMultiobjective} proposed to find robust architectures against multiple $l_{\infty}$ norm attacks by using evolutionary algorithms. In general, the robustness evaluation metrics are indirect during differentiable NAS, while that in non-differentiable NAS can be direct, such as the robust accuracy of models under adversarial attacks.

\begin{figure*}[t]
	\begin{center}
		\subfigure[Clean]{
			\includegraphics[width=.15\textwidth]{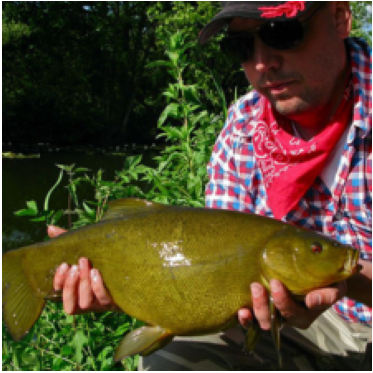}
			\label{a}
		}
		\subfigure[Hue]{
			\includegraphics[width=.15\textwidth]{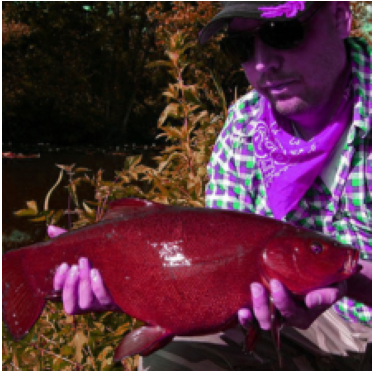}
			\label{b}
		}
		\subfigure[Saturation]{
			\includegraphics[width=.15\textwidth]{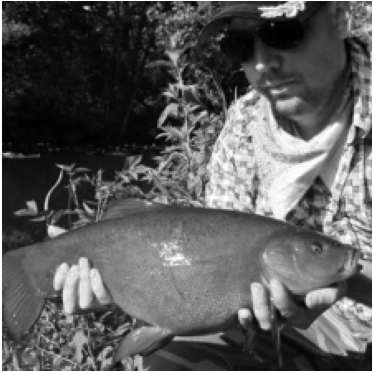}
			\label{c}
		}
		\subfigure[Rotation]{
			\includegraphics[width=.15\textwidth]{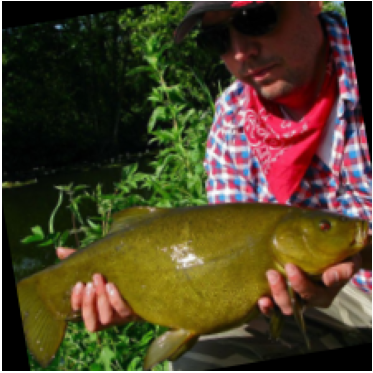}
			\label{d}
		}
		\subfigure[Brightness]{
			\includegraphics[width=.15\textwidth]{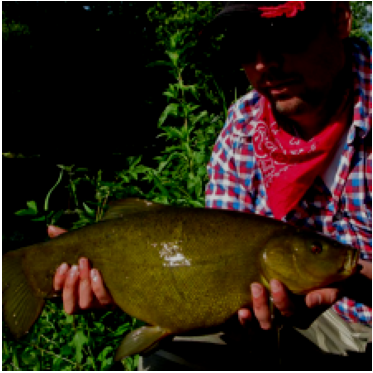}
			\label{e}
		}
		\subfigure[Contrast]{
			\includegraphics[width=.15\textwidth]{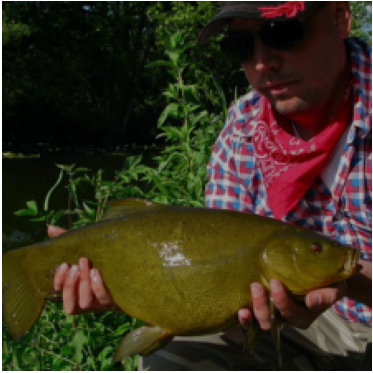}
			\label{f}
		}

		\subfigure[FGSM-$\mathscr{l}_{\infty}$]{
			\includegraphics[width=.15\textwidth]{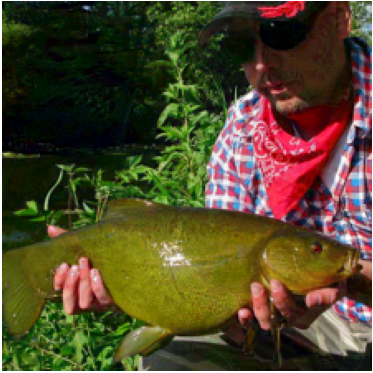}
			\label{f}
		}
		\subfigure[PGD-$\mathscr{l}_{\infty}$]{
			\includegraphics[width=.15\textwidth]{5pgd.png}
			\label{f}
		}
		\subfigure[MI Attack-$\mathscr{l}_{\infty}$]{
			\includegraphics[width=.15\textwidth]{5pgd.png}
			\label{g}
		}
		\subfigure[PGD-$\mathscr{l}_{2}$]{
			\includegraphics[width=.15\textwidth]{5pgd.png}
			\label{h}
		}
		\subfigure[MI Attack--$\mathscr{l}_{2}$]{
			\includegraphics[width=.15\textwidth]{5pgd.png}
			\label{i}
		}
		\subfigure[CAA]{
			\includegraphics[width=.15\textwidth]{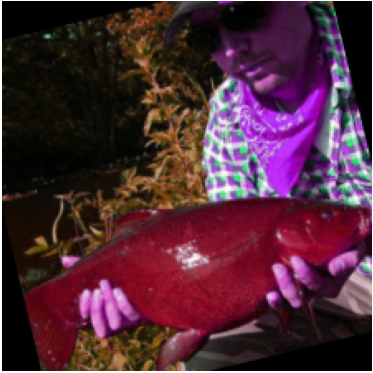}
			\label{j}
		}
		\caption{The illustration of adversarial examples generated by different attack methods. }
		\label{panda}
	\end{center}
\end{figure*}

Though many works above-mentioned have been developed to find more robust neural architectures by NAS techniques, they only focus on improving the robustness under $l_{\infty}$ norm-based adversarial attacks. It remains an open problem to find comprehensively robust architectures under multiple types of evaluations, including different types of $l_{p}$ $(p=2,\infty)$ norm-based adversarial attacks, semantic adversarial attacks \cite{joshi2019semantic,hosseini2018semantic,bhattad2019unrestricted} and composite adversarial attack (CAA) \cite{hsiung2023caa}. An illustration of different types of attacks is presented in Figure \ref{panda}. When comprehensively robust neural architectures are hoped to be found by NAS, the largest challenge is the huge computational cost of robustness performance evaluation. The robustness evaluation under adversarial attack is farther slower than the standard accuracy, which is also called clean accuracy. In addition, increasing the type of robustness evaluation would also further increase the computational cost. These problems will incur that even though the surrogate model technique \cite{sobester2008engineering} is applied to predict the robustness performance of neural architectures, preparing the training data consisting of hundreds of data pairs still needs to take a huge computational burden.

To address the above limitations, this work introduces a novel approach for discovering comprehensively robust neural network architectures using a multi-fidelity surrogate-based evolutionary algorithm that incorporates multiple types of robustness evaluations. The key contributions of this work can be succinctly stated as follows:
\begin{itemize}
	\item We first propose to search for comprehensively robust architectures under multiple types of evaluations using the weight-sharing-based NAS method, including different $l_{p}$ norm attacks, semantic adversarial attacks, and composite adversarial attacks.
	
	\item We reduce the number of robustness evaluations by the correlation analysis, which can incorporate similar evaluations and decrease the evaluation cost.

	\item We propose to decrease the search cost through multi-fidelity evaluations. A multi-fidelity online surrogate during optimization is constructed. On the basis of the surrogate constructed by low-fidelity data, the online high-fidelity data is utilized to finetune the surrogate.

\end{itemize}

 The paper is structured as follows: In Section \ref{sec2}, a brief introduction will be provided on the related techniques used in this work. Section \ref{sec3} will elaborate on the proposed approach to search for comprehensively robust neural architectures in detail, including encoding, correlation analysis for different evaluations, the multi-fidelity surrogate, and the search framework based on the archive. Experimental settings and results are presented in Section \ref{sec4}. Finally, we make conclusions and discuss future work in Section \ref{sec5}.

\section{Background}\label{sec2}
This section introduces the background of adversarial attacks and NAS for robust neural architectures.

\subsection{Adversarial Attacks}
The goal of adversarial attacks is to generate specific perturbations to make DNN models output wrong results. From the magnitude of the generated perturbations, existing methods can be categorized into $l_{p}$ norm-based and semantic adversarial attacks. In addition, composite adversarial attacks that combine multiple types of adversarial attacks together are also developed. In this section, we introduce the related work about $l_{p}$ norm adversarial attack, semantic adversarial attack, and composite adversarial attack, respectively.
\subsubsection{$\mathscr{l}_{p}$-norm Adversarial Attacks}\label{watk}

The $l_{p}$ norm-based adversarial attack is an optimization problem with the constraint intrinsically, which can be formulated as Eq. \ref{attack}.
\begin{equation}
\begin{array}{l}
\underset{\Delta x}{\arg \max }  \mathcal{L}\left(x_{adv}, y ; \mathcal{F}\right)\\
x_{a d v}=x+\Delta x \\
\text { s.t. }\|\Delta x\|_{p} \leq \varepsilon
\end{array}
\label{attack}
\end{equation}
where $p$ represents the norm type of the perturbation such as $l_{2}$ and $l_{\infty}$. $\mathcal{L}$ is the loss function. $x$ is the original image, which is also called the clean image. $\Delta x$ stands for the perturbation, whose magnitude is $\epsilon$. $\mathcal{F}$ is the DNN model. The true label is denoted as $y$. $x_{adv}$ is the final generated adversarial example.

To solve Eq. \ref{attack}, many adversarial attack algorithms have been developed. The fast gradient sign method (FGSM) is the earliest adversarial attack method, which was proposed by Goodfellow et al. \cite{goodfellow2014explaining}, as shown in Eq. \ref{fgsm}.

\begin{equation}
x_{a d v}=\operatorname{clip}_{[0,1]}\left\{x+\epsilon \cdot \operatorname{sign}\left(\nabla_{x} \mathcal{L}(x, y ; \mathcal{F})\right)\right\}
\label{fgsm}
\end{equation}

FGSM utilized the single-step updating strategy, which is easily trapped in the local optimum. To solve the problem, Madry et al. \cite{pgd2018} further developed the multi-step updating method, PGD, which is presented in Eq. \ref{pgd}.
\begin{equation}
x_{l+1}=\operatorname{project}\left\{x_{l}+\epsilon_{step} \cdot \operatorname{sign}\left(\nabla_{x} \mathcal{L}\left(x_{l}, y ; \mathcal{F}\right)\right)\right\}
\label{pgd}
\end{equation}
where $\epsilon_{step}$ denotes the magnitude of perturbation at each iteration. In the above-mentioned process, the initial image can either be the original image or a perturbed image resulting from random initialization. 

Carlini and Wagner \cite{carlini2017towards} proposed a modified loss function for adversarial attacks. This loss function calculates the distance between the logit values of the target label $y$ and the second most likely class, as shown in Eq. \ref{cw}.

\begin{equation}
\mathcal{L}_{C W}(x, y ; \mathcal{F})=\max \left(\max _{i \neq y}\left(\mathcal{F}(x)_{(i)}\right)-\mathcal{F}(x)_{(y)},-\kappa\right)
\label{cw}
\end{equation}
where the parameter $\kappa$ serves to define the margin between the logit value of the adversarial class and the logit value of the second most likely class \cite{Composite2021}.

In addition to the adversarial attacks previously described, other works such as MultiTargeted (MT) attack \cite{gowal2019alternative}, Momentum Iterative
(MI) attack \cite{dong2018boosting}, and Decoupled Direction and Norm (DDN) attack \cite{rony2019decoupling} are also further developed. These methods have been designed to further improve the performance of adversarial attacks by changing the way gradients are updated, the magnitude of perturbations, the number of attack iterations, and the loss functions used. 
\subsubsection{Semantic Adversarial Attacks}\label{batk}

For most of the semantic perturbations, the
parameters that need to be optimized are continuous. Under the white-box attack setting, 
the parameters of semantic attacks can be updated by the gradient descent algorithm. Specifically, the paragraph outlines how the parameters can be updated for five different types of semantic perturbations, namely hue, saturation, brightness, contrast, and rotation. To optimize these perturbations, the iterative gradient sign method is extended for $T$ iterations, which is defined as:

\begin{equation}
\delta_k^{t+1}=\operatorname{clip}_{\epsilon_k}\left(\delta_k^t+\alpha \cdot \operatorname{sign}\left(\nabla_{\delta_k^t} J\left(\mathcal{F}\left(A_k\left(X ; \delta_k^t\right)\right), y\right)\right)\right),
\label{sgd}
\end{equation}
where $t$ is the iteration number, $\mathcal{F}$ stands for the DNN model, $X$ denotes the input images, $y$ is the prediction label, $\epsilon_k$ is the perturbation interval. Assume $\epsilon_k = [\alpha_k, \beta_k]$, the element-wise clipping
operation $\operatorname{clip}_{\epsilon_k}$ is expressed as:

\begin{equation}
\operatorname{clip}_{\epsilon_k}(z)=\operatorname{clip}_{\left[\alpha_k, \beta_k\right]}(z)=\left\{\begin{aligned}
\alpha_k & \text { if } z<\alpha_k, \\
z & \text { if } \alpha_k \leq z \leq \beta_k, \\
\beta_k & \text { if } \beta_k<z
\end{aligned}\right.
\label{handle}
\end{equation}

The following description provides an explanation of each semantic attack.

\textbf{Hue}: The Hue attack can transfer clean images from RGB space to hue-saturation-value (HSV) space, causing a dropping in accuracy. The Hue value ranges from 0 to 2$\pi$. Hence, the maximum perturbation interval of Hue attack is [-2$\pi$,2$\pi$] \cite{hsiung2022carben}. 

\begin{equation}
x_H^t=\operatorname{Hue}\left(x_{\mathrm{adv}}^t\right)=\operatorname{clip}_{[0,2 \pi]}\left(x_H+\delta_H^t\right) 
\label{hue}
\end{equation}

\textbf{Saturation}: The Saturation attack can change the colorfulness of clean images by modifying the Saturation value, which ranges from 0 to 1. If the saturation value tends to be 1, the image becomes more colorful, while that is a gray-scale image if the saturation value is 0 \cite{hsiung2023caa}.

\begin{equation}
x_S^t=\operatorname{Sat}\left(x_{\text {adv }}^t\right)=\operatorname{clip}_{[0,1]}\left(x_S \cdot \delta_S^t\right)  
\label{saturation}
\end{equation}

\textbf{Brightness and Contrast}: Brightness and contrast are different from hue and saturation in that they are defined in the RGB color space (pixel space) and determine the brightness and darkness differences of images. In our implementation, we first convert the images from the [0, 255] scale to the [0, 1] scale. The perturbation interval for brightness is defined as
$\epsilon_B = [\alpha_B, \beta_B]$, where $-1 \leq \alpha_B \leq \beta_B \leq 1$, while for contrast it is defined as   $\epsilon_C = [\alpha_C, \beta_C]$, where $-1 \leq \alpha_C \leq \beta_C \leq 1$.

Similarly to the hue and saturation attacks, we choose initial perturbations $\delta^{0}_{B}$ and $\delta^{0}_{C}$ uniformly from  $\epsilon_B$ and $\epsilon_C$, respectively, and update them using equation \ref{sgd}. The perturbed image $x^{t}_{adv}$ under the brightness attack is then obtained by adding the perturbation to the original image x, clamping the resulting values to be between 0 and 1, and scaling the result back to the [0, 255] scale.
The Brightness and Contrast attacks determine the lightness, darkness, and brightness difference of images in RGB color space \cite{hsiung2022carben}, which can be obtained by Eq. \ref{brightness}.

\begin{equation}
x_{\mathrm{adv}}^t=\operatorname{clip}_{[0,1]}\left(x+\delta_B^t\right) \text { and } x_{\mathrm{adv}}^t=\operatorname{clip}_{[0,1]}\left(x \cdot \delta_C^t\right) 
\label{brightness}
\end{equation}

\textbf{Rotation}: The purpose of this transformation is to discover an angle of rotation that maximizes the loss of the rotated image. The rotation algorithm was developed by \cite{riba2020kornia}. Assuming we have a square image $x$, we can denote a pixel's position as $\left(i,j\right)$, and the center of $x$ as $\left(c,c\right)$. To calculate the new position $\left(i',j'\right)$ of a pixel rotated by an angle of $\theta$ degrees, we can use the following formula:

\begin{equation}
\left[\begin{array}{l}
i^{\prime} \\
j^{\prime}
\end{array}\right]=\left[\begin{array}{c}
\cos \theta \cdot i+\sin \theta \cdot j+(1-\cos \theta) \cdot c-\sin \theta \cdot c \\
-\sin \theta \cdot i+\cos \theta \cdot j+\sin \theta \cdot c+(1-\cos \theta) \cdot c
\end{array}\right]
\label{Rotation}
\end{equation}

\subsubsection{Composite Adversarial Attacks}\label{caa}
\begin{figure}[h]
	\centering
	\includegraphics[scale=0.5]{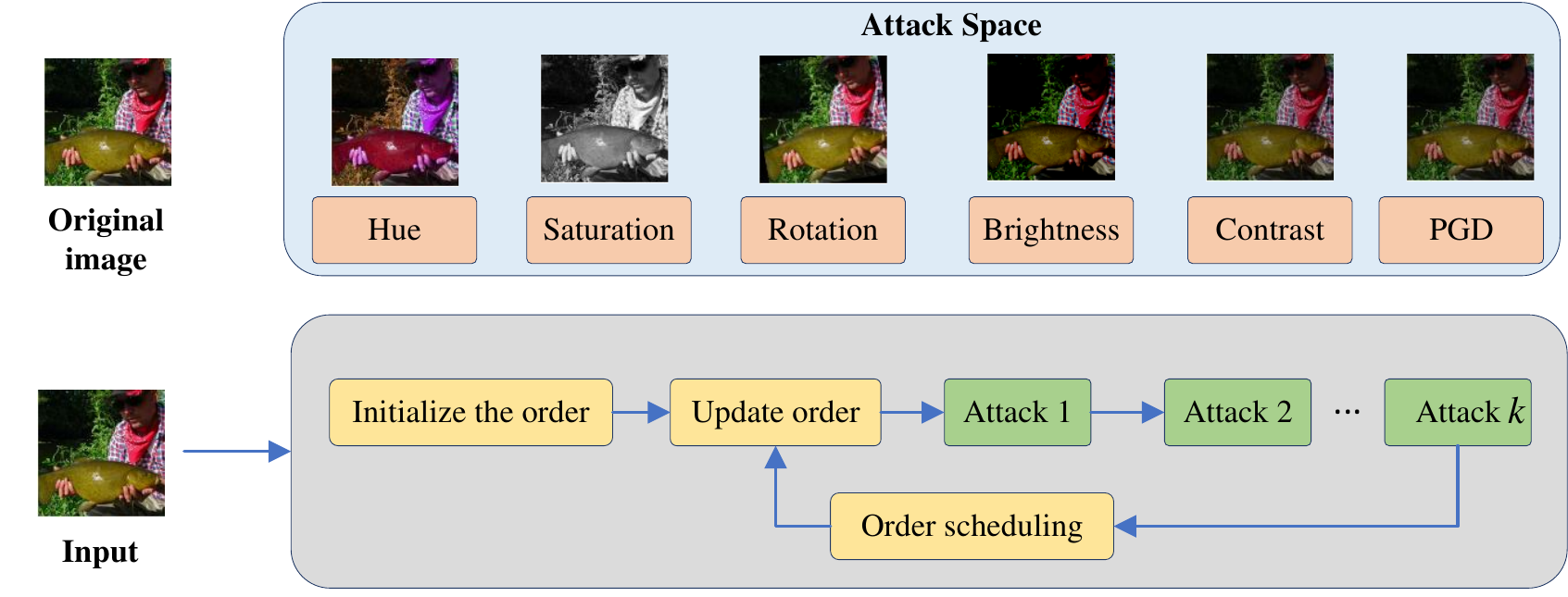}
	\caption{The illustration of composite adversarial attack.}
	
	\label{csp_process}
\end{figure}

As for the white-box model, each attack above-mentioned can obtain the optimal perturbations by iteratively updating according to the gradient of model predictions. CAA constructed the mathematical model of combining multiple types of attacks, which includes two modes, namely the fixed and scheduled. The fixed one means that the ensemble attack with the fixed sequence is utilized. The scheduled one means that the order of the attack sequence can be adaptively adjusted with the batch data, achieving a higher attack success rate. An illustration of the process of CAA is presented in Figure. \ref{csp_process}. The attack space includes five types of semantic adversarial attacks. When the attack sequence is given, the input image would be fed to different attacks subsequently. If the scheduled mode is selected, the order will be updated to complete the semantic attack.

\subsection{NAS for Robust Architectures}
Neural Architecture Search (NAS) is one technique that belongs to auto machine learning (AutoML), whose objective is to automatically discover high-performing neural network architectures from a given search space. To accomplish this, various effective search strategies, such as reinforcement learning, evolutionary algorithms, and Bayesian optimization, are utilized. NAS methods can search architectures without much domain knowledge, further improving the performance of DNN with the predefined evaluation metrics. 

In previous works, much effort has been devoted to improving the clean accuracy of DNN models. With the development of adversarial attack and defense community, finding robust architectures has emerged as one increasing direction. Thus over the past years, some works tried to apply NAS techniques to find robust architectures against AEs. To realize that, the robustness evaluation metrics are first needed to be determined. The efficient search strategies can be combined to find robust architectures within the given search space. In general, existing NAS methods for robust architectures
can be categorized into differentiable and non-differentiable
based. The non-differentiable-based methods
evaluate the robustness of each architecture using adversarial
attacks like PGD during the search process. The processes of searching and evaluating the architectures are separate,
which is relatively time-consuming. Guo et al. \cite{whencvpr} applied
the weight-sharing-based random search to find robust network architectures
toward PGD. There are also some works that utilize evolutionary algorithms to find more
robust architectures \cite{vargas2019evolving,liu2021multi,liu2022bi,ning2020discovering}. Chen et al. \cite{chen2020anti} applied reinforcement learning algorithms to find robust neural architectures. In contrast, the differentiable methods
adopt the differentiable metric into differentiable architecture
search methods, which greatly accelerates the search process.
Hosseini et al. \cite{Differentiable2021} proposed DSRNA, which combines the
quantified metrics, such as the norm of the Jacobian matrix and certified lower bound, into the differential architecture search to
efficiently find robust model architectures. Mok et al. \cite{AdvRush2021}
proposed AdvRush, which finds more robust architectures
based on the $F$ norm of the Hessian matrix. Qian et al. \cite{qian2022robust} defined the similarity between the adversarial examples and clean examples in the feature space as the robustness metric to guide the differentiable architecture search, which also realizes the search of robust architectures. Their method is termed RNAS. Ou et al. \cite{ou2022differentiable} developed the two-stage training method in differentiable architecture search in their carefully crafted search space. Dong et al. \cite{dong2020adversarially} proposed RACL, which combines the Lipschitz constant into the differentiable NAS method to find robust architectures. Though these above-mentioned works have been developed in the community of NAS for robust architectures, the related work that finds comprehensively more robust architectures has not been explored due to the higher computational burden.

\section{Proposed Method}\label{sec3}

Deep neural networks (DNNs) have proved to be vulnerable to adversarial attacks, and the robustness performance of a model can vary depending on the type of adversarial attack. Thus to find a comprehensively robust neural architecture, the robust accuracy of models under multiple types of attacks is needed to be considered. Unfortunately, evaluating the robustness using all types of adversarial attacks can lead to a huge computational burden. To address the above challenges, we propose an algorithm to search for robust architectures, termed ES-CRNA-ME in short. ES-CRNA-ME is designed to search for neural architectures that are less sensitive to different types of adversarial attacks. 

\begin{figure*}[t]
	\centering
	\includegraphics[width=.95\textwidth]{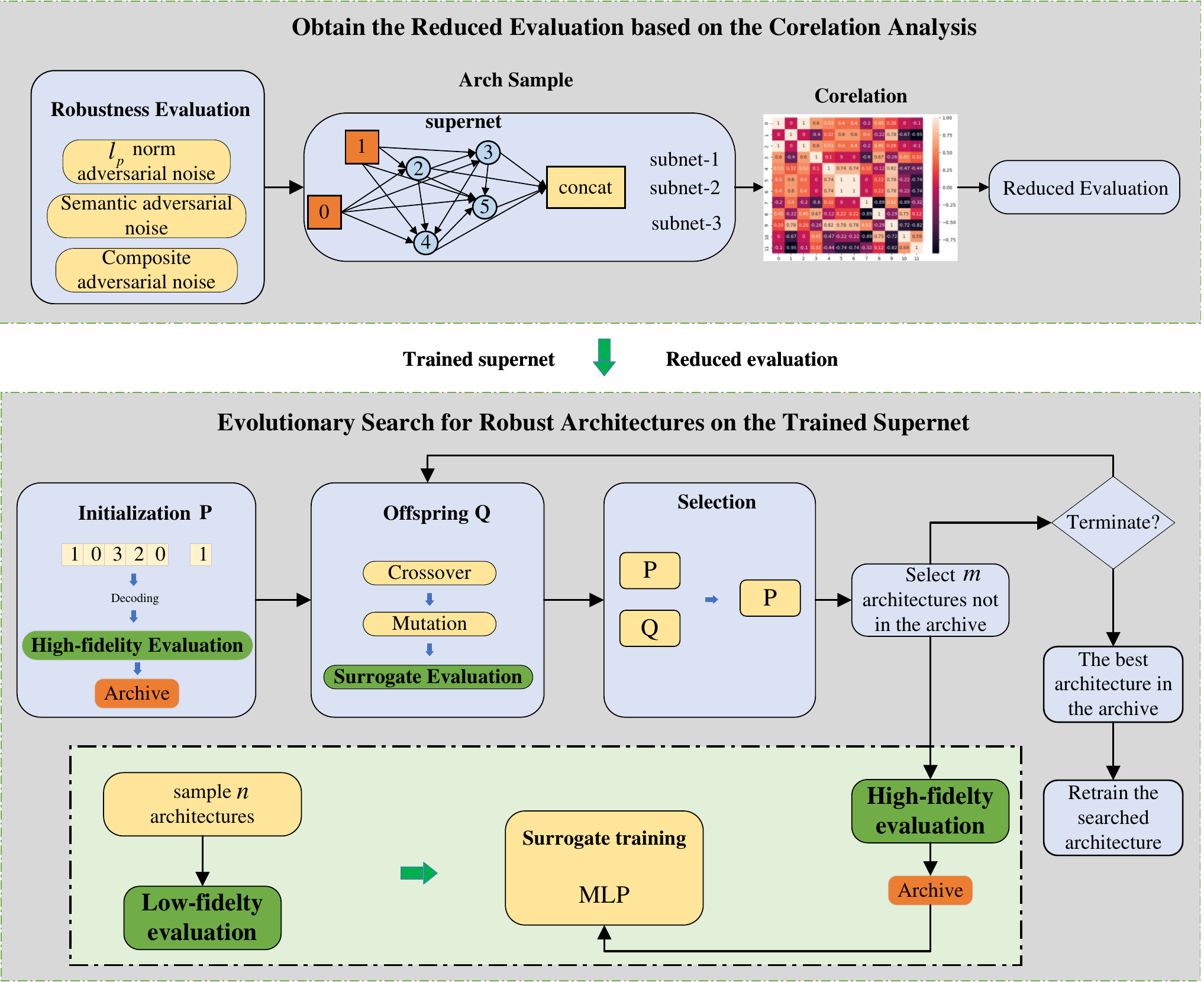}
	\caption{The framework of the proposed efficient search of comprehensively robust neural architectures via multi-fidelity evaluation.}
	\label{framework}
\end{figure*}

In the following, we will elaborate on the main components of the proposed algorithm. We start with the encoding strategy, followed by a description of the correlation analysis for reducing the type of robustness evaluations. Then the multi-fidelity surrogate model in our proposed method is introduced. Lastly, we will describe the whole framework of the proposed method.

\subsection{Encoding}\label{3}

\begin{figure*}[t]
	\centering
	\includegraphics[width=.9\textwidth]{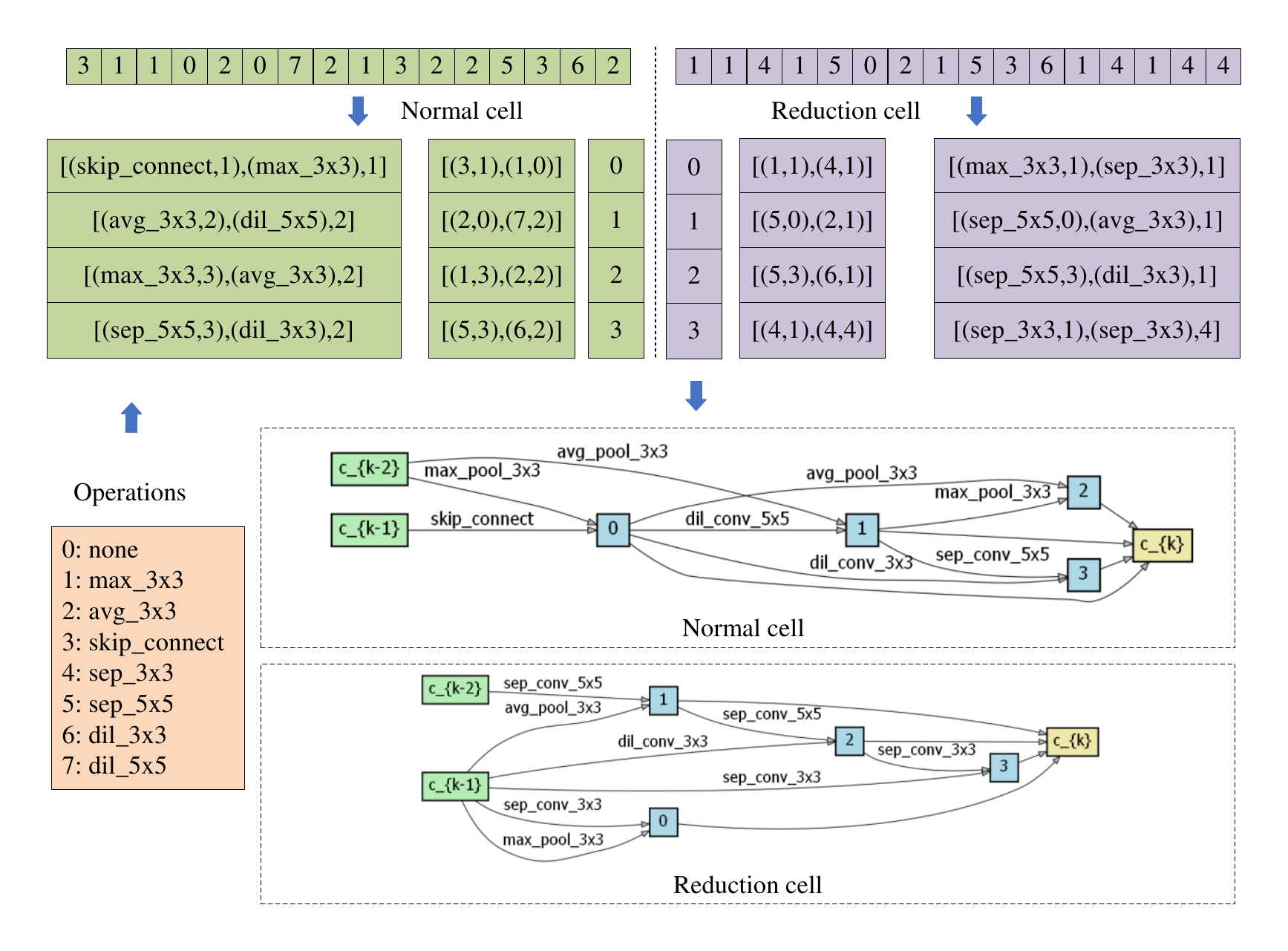}
	\caption{The encode strategy in our proposed method.}
	\label{encode}
\end{figure*}

This work employs the cell-based search space proposed by Zoph et al. \cite{zoph2018learning} consisting of normal and reduction cells. Normal cells do not change the size of the feature map, while reduction cells reduce that half. In this space, each cell consists of seven nodes, where the first two nodes are inputs from the two previous cells, which are denoted as $c$$\_${k-2} and $c$$\_${k-1}, respectively. The number of intermediate nodes is four, which are denoted as 0, 1, 2, and 3. Two edges are allowed to connect between any two nodes. Each edge can take one of several predefined operations. At last, the outputs of four intermediate nodes will be combined as inputs and enter the seventh node together, whose output is denoted as $c$$\_${k}, which is also the output of the current cell. In this way, the whole neural architecture can be represented by a connected directed acyclic graph (DAG). To encode this type of architectures, the adopted encoding method is the same as that proposed by Liu et al. \cite{liu2021multi} in previous work. In normal cells and reduction cells, each node receives information from two other nodes. For a given node, we need to determine which node the inputs come from and which operation operator to select for the edge between the two nodes. Thus, four variables can determine the input configuration of a node. Because there are eight nodes in two types of cells in total, we can use the 32-dimensional decision variables to represent a neural architecture. For example, the sequence that represents each cell is divided into four segments. Each segment has two tuples. In each tuple, the first bit indicates the operation to be performed, and the second bit indicates which nodes this node is connected to. A normal cell, for instance, is encoded by the segment ‘[(3, 1), (1, 0)]’, indicating that nodes 1 and 0 have operations ‘skip$\_$connect’ and ‘max$\_$3x3’ applied to them, respectively. In the encoding, two input nodes are denoted as 0 and 1. An illustration of encoding this type of architecture can be seen in Figure \ref{encode}. The search space includes eight operations, namely none, max$\_$pool$\_$3x3, avg$\_$pool$\_$3x3, \\sep$\_$conv$\_$3x3,  dil$\_$conv$\_$3x3, skip$\_$connect, sep$\_$conv$\_$5x5, and sep$\_$conv$\_$5x5. In Figure \ref{encode}, we simplify some operations to easier representations.

\subsection{Correlation analysis for different evaluations}\label{3.1}
The core idea of weight-sharing-based architecture search is first to construct the supernet consisting of all possible subpaths. Each subpath represents a sub-model. After the supernet is trained, the evolutionary algorithm can be easily combined to find the near-optimal architecture. In this process, the robustness evaluation on the trained supernet is needed to be performed. Though weight-sharing-based architecture search can decrease the search cost greatly, it does not solve the computational burden due to the consideration of searching for comprehensively robust architectures. In particular, with the increasing number of the type of evaluations, the computational cost of robustness evaluation can further increase.

To alleviate this problem, in this work, we propose to utilize correlation analysis to reduce the number of robustness evaluations by incorporating similar evaluations.
Assume that there exist $n$ evaluations, which are expressed as   
\begin{equation}
F=\{{{f}_{1}},{{f}_{2}},\cdots ,{{f}_{n}}\}
\end{equation}
where $f$ is the function to calculate the robust accuracy under the adversarial attack. The robust accuracy is calculated as
\begin{equation}
Robust\text{ }Accuracy\text{ =}\frac{{{n}_{adv}}}{{{n}_{total}}}
\label{ra_cal}
\end{equation}
where ${n}_{total}$ stands for the number of total test samples that are generated by the adversarial attack. ${n}_{adv}$ denotes the number of examples that are predicted rightly. In this paper, 11 adversarial attacks are considered. Thus, 11 robust accuracies are calculated. Besides, the clean accuracy of the model is also taken as one of the objectives, which is defined as the accuracy of the model on the clean examples that are not perturbed by any adversarial attack. Hence, there exist 12 evaluations of the accuracy of the models in total.

Then we sample some architectures from the supernet and evaluate their robust accuracies. If two types of evaluation are similar, we will merge them together. The similarity adopts the correlation value as a metric. In this way, $n$ evaluations can be further reduced to $m$ evaluations.
\begin{equation}
\tilde{F}=\{{{\tilde{f}}_{1}},{{\tilde{f}}_{2}},\cdots ,{{\tilde{f}}_{m}}\}\
\end{equation}

The robust accuracy of models is estimated as Eq. \ref{comprehensive}.
\begin{equation}
RA=\frac{{{k}_{1}}{{{\tilde{f}}}_{1}}+{{k}_{2}}{{{\tilde{f}}}_{2}}+\cdots +{{k}_{m}}{{{\tilde{f}}}_{m}}}{n}
\label{comprehensive}
\end{equation}
Assume the accuracy from ${{\tilde{f}}}_{m}$ to ${{\tilde{f}}}_{n}$ can be represented by ${{\tilde{f}}}_{m}$. The coefficient ${k}_{m}$ of ${{{\tilde{f}}}_{m}}$ of Eq. \ref{comprehensive} is calculated as 
\begin{equation}
{{k}_{m}}=\frac{{{f}_{m}}+{{f}_{m+1}}+\cdots +{{f}_{n}}}{{{f}_{m}}}\
\label{coefficient}
\end{equation}
Other coefficients, including ${{k}_{i}} (i=1,2,...,m)$, can also be determined by this way. Therefore, we just need to calculate the robust accuracy under $m$ adversarial attacks rather than the original $n$ attacks.

\subsection{Multi-fidelity surrogate}\label{obj}
Though the above-mentioned correlation analysis can decrease the evaluation cost to some extent, there still exist some evaluations under multiple types of attacks that need to be performed. To further reduce the search cost, we propose the multi-fidelity strategy that learns the mapping from the architectures to the corresponding comprehensively robust accuracy. To realize that, we use the Multilayer Perceptron (MLP) as the surrogate model. We first train the surrogate using low-fidelity data. The low-fidelity evaluation means that the robust accuracy of models is evaluated using a small number of data. In contrast, the high-fidelity evaluation uses a large number of data to calculate the robust accuracy of models. Hence, the high-fidelity evaluation is more accurate. However, it is very time-consuming to train the surrogate using a large number of high-fidelity evaluations. Thus, the low-fidelity data can provide a good initialization for training the surrogate. Then we use the high-fidelity data to update the surrogate iteratively.

\subsection{The search framework based on the archive}

Combined with the above two strategies to decrease the search cost, we introduce the search framework based on the archive in the following. The whole process of our proposed method is presented in Figure \ref{framework}. The corresponding pseudocode can be seen in Algorithm \ref{alg1}.

\begin{algorithm}[htbp]\footnotesize{
		\caption{The procedure of the proposed ES-CRNA-ME} \algblock{Begin}{End}
		\label{alg1}
		\textbf{Input:} The population size $N$, the maximal generation number $G$, the trained supernet, different types of adversarial attacks \\
		\textbf{Output:} The near-optimal trained architecture
		\begin{algorithmic}[1]
			
			\State Sample $m$ architectures from the supernet and evaluate their robust accuracies using the supernet on different types of attacks
			\State Calculate the correlation between different attacks and merge the similar attacks according to Eq. \ref{comprehensive}
			\State // \textbf{Train the surrogate using low-fidelity data}
			\State Sample $n$ architectures from the supernet and evaluate their robust accuracy using the low-fidelity data, then train the surrogate 
			\State Initialize the population $P$ with $N$ individuals by randomly generating different architectures
			\State \textbf{High-fidelity evaluation:} Evaluate the robust accuracy of population $P$ using high-fidelity data and preserve the results into the archive
			\For {$t=0$ to $G$}
			\State \textbf{Crossover and mutation:} Generate $N$ offspring individuals through crossover and mutation operations, and all the individuals are evaluated using the trained surrogate
			\State \textbf{Selection:} Select $N$ individuals to form the new population and $m$ individuals that are not in the archive to perform the next high-fidelity evaluation
			\State Evaluate these $m$ individuals using high-fidelity data
			\State // \textbf{Update the surrogate using high-fidelity data}
			\State Using the high-fidelity data finetune the trained surrogate model
			\EndFor
			\State Obtain the best individual in the archive
			\State \textbf{Final training:} Decode the best individual from the archive for the final deep training
			\State \Return The trained architecture
	\end{algorithmic}}
\end{algorithm}

As presented in Figure \ref{framework} and Algorithm \ref{alg1}, ES-CRNA-ME starts with determining the type of robustness evaluations according to the trained supernet, which is described in Section \ref{3.1}. Some architectures are sampled from the supernet and evaluated under different types of robustness evaluations. The correlation analysis is performed to reduce the number of evaluations. Then we use the low-fidelity data to train the surrogate model, which can substitute the evaluation during evolutionary optimization for robust architectures.

Then based on the reduced robustness evaluation, the trained supernet, and the surrogate model, we conduct the search for comprehensively robust architectures. At first, the initial population with $n$ individuals is randomly generated, which is all evaluated by high-fidelity data. We use an archive to preserve high-fidelity data. In addition, all the high-fidelity data is utilized to finetune the surrogate model. During the optimization, we generate the offspring by the crossover and mutation operations iteratively. When we conduct the selection operation, the evaluation of all individuals is based on the prediction of the surrogate model. In this procedure, to guarantee population diversity, if the selected individuals exist in the archive, we will regenerate the individuals randomly. After $G$ generations, the best individual in the archive will be output as the final searched architecture. At last, we retrain the searched architecture and obtain a comprehensively robust model.

\begin{table}[htbp]
	\caption{The settings of different attacks.}
	\label{attacksetting}
	\centering
	\begin{tabular}{ccc}
		\hline
		Methods & Magnitude & Iteration number\\ \hline
		Hue      & [-$\pi$, $\pi$]    &1 \\
		Saturation&[0.7,1.3]&1\\
		Rotation            & [-10, 10]   & 1\\
		Brightness            & [-0.2, 0.2]   &1 \\
		Contrast            & [0.7, 1.3]    & 1\\
		CAA            & -    & 1\\
		FGSM-$\mathscr{l}_{\infty}$            & 1/255    & 7\\
		PGD-$\mathscr{l}_{\infty}$&  1/255     &7 \\
		MI-attack-$\mathscr{l}_{2}$, &  1/255     &7 \\
		PGD-$\mathscr{l}_{2}$ &  1/255     &7 \\
		MI-attack-$\mathscr{l}_{2}$&  1/255     & 7\\
		\hline
	\end{tabular}
\end{table}

\begin{table*}[htbp]
	\caption{The performance of different architectures on CIFAR10. The first four models are manually designed architectures, while the last seven models are the searched cell-based architectures by NAS. Among that, ES-CRNA-ME is the searched architectures by us. The best results are highlighted in bold. }
	\label{Res10}
	\centering
	\resizebox{\textwidth}{!}{
		\begin{tabular}{cccccccccccccc}
			\hline
			Networks    & Clean & Hue & Saturation   & Rotation  & Brightness& Contrast&CAA& FGSM-$\mathscr{l}_{\infty}$& PGD-$\mathscr{l}_{\infty}$&MI attack-$\mathscr{l}_{2}$& PGD-$\mathscr{l}_{2}$&MI attack-$\mathscr{l}_{2}$& Avg\\ 
			
			\hline
			
			GoogleNet&\textbf{94.47}&81.21&92.61&86.66&89.15&91.49&4.50&44.2&28.0&27.4&65.6&65.4&64.22\\
			MobileNetV2 & 93.01&77.28 &91.32&84.02&86.88&89.40&4.47&46.4&25.8&25.2&62.8&62.2&62.40\\
			DenseNet121&94.34&80.67&93.10&87.00&89.10&91.50&6.07&55.2&34.6&33.4&67.6&66.0&66.55\\
			VGG19&92.34&75.73&90.11&83.15&86.39&88.32&8.84&60.0&47.0&46.4&75.6&75.4&69.11\\
			\hline
			
			Random &87.99&66.59&85.00&75.70&79.71&82.90&3.78&38.6&28.0&27.6&59.8&59.6&51.08\\ 
			DARTS & 94.14 & 82.60 & 92.77 & \textbf{87.04} & \textbf{89.79} &\textbf{91.64}    & 6.82&54.8&34.2&33.2&71.0  &69.0  &67.25 \\
			PDARTS&93.78&82.29&92.51&86.49&89.32&91.63&7.95&57.4&43.0&42.8&73.4&72.6&69.43\\
			RACL&93.31&82.26&92.21&85.22&89.20&90.87&7.91&53.4&35.0&35.0&70.6&69.4&67.04\\
			RNAS&91.61&78.28&90.14&84.06&87.05&88.99&11.44&63.0&52.2&52.4&\textbf{78.6}&\textbf{78.6}&71.36\\
			Advrush & 93.85 & \textbf{83.26} & 92.76 & 86.70 & 89.58 & 91.57    & 8.62&55.6&37.6& 37.6  &73.2   &71.8&68.51 \\
			ES-CRNA-ME &92.31&79.67&90.62&84.90&87.75&89.61&\textbf{12.5}&\textbf{64.4}&\textbf{52.4}&\textbf{52.4}&77.4&77.0& \textbf{71.75}\\
			\hline
	\end{tabular}}
\end{table*}

\begin{table*}[htbp]
	\caption{The performance of different architectures on CIFAR100. }
	\label{results100}
	\centering
	\resizebox{\textwidth}{!}{
		\begin{tabular}{cccccccccccccc}
			\hline
			Networks    & Clean & Hue & Saturation   & Rotation  & Brightness& Contrast&CAA& FGSM-$\mathscr{l}_{\infty}$& PGD-$\mathscr{l}_{\infty}$&MI attack-$\mathscr{l}_{2}$& PGD-$\mathscr{l}_{2}$&MI attack-$\mathscr{l}_{2}$& Avg\\ \hline
			
			GoogleNet&76.26&39.56&70.86&62.60&64.22&68.13&0.99&20.6&7.00&7.00&33.0&30.8&33.00\\
			MobileNetV2 & 72.00&36.09&65.41&56.37&59.07&62.82&1.27&21.8&6.4&6.4&30.8&30.4&37.40\\
			DenseNet121&\textbf{77.04}&40.20&71.07&62.96&64.41&68.34&1.38&25.2&12.4&11.8&36.0&33.0&41.98\\
			VGG&67.73&29.25&60.25&52.28&55.48&58.42&1.04&24.8&14.8&14.2&40.2&39.6&38.17\\
			\hline
			
			Random &42.58&17.47&34.28&32.01&31.35&33.16&1.02&17.6&13.4&13.2&25.6&25.0&23.82\\ 
			DARTS & 75.85&40.00&70.13&63.57&64.41&68.15&2.00&33.2&18.8&18.4 &43.2&41.4&44.93\\
			PDARTS&76.27&\textbf{40.22}&\textbf{71.07}&\textbf{64.19}&\textbf{65.86}&\textbf{69.39}&2.21&33.4&20.0&20.8&46.2&46.0&46.30\\
			RACL&73.99&38.59&68.39&61.36&62.87&66.50&2.03&31.0&19.2&19.0&40.8&39.2&43.58\\
			RNAS&68.58&33.17&62.17&57.10&57.80&60.99&2.20&\textbf{35.6}&24.8&24.8&48.6&48.0&43.65\\
			Advrush & 76.04&39.62&70.78&63.87&65.45&68.68&2.10&31.4&18.6& 17.8&42.2&42.2&44.90\\
			ES-CRNA-ME&73.29&36.22&67.84&61.22&61.59&65.84&\textbf{2.26}&\textbf{35.6}&\textbf{26.0}&\textbf{25.8}&\textbf{51.4}&\textbf{50.6}&\textbf{46.47}\\
			\hline
	\end{tabular}}
\end{table*}

\section{Experiments}\label{sec4}
To show the effectiveness of ES-CRNA-ME, we conduct experiments on the CIFAR10 and CIFAR100 datasets. In section \ref{41}, we give a brief description of the experiment protocol. In section \ref{42} and section \ref{43}, the experimental results on CIFAR10 and CIFAR100 datasets are presented, respectively. In section \ref{44}, we make the analysis of the relativity of different evaluations. In section \ref{45}, we analyze the performance of the multi-fidelity surrogate model and the efficiency of the two strategies. Finally, in section \ref{46}, we discuss the strength and weaknesses of our method.
\subsection{Experiment Protocol}\label{41} 

Both CIFAR10 and CIFAR100 datasets include 50,000 training images and 10,000 test images. In our setting, the evaluation results on 100 images are addressed as the low-fidelity evaluation, while that on 5,000 images stands for the high-fidelity evaluation. The batch size is set to 32. The number of sampled architectures using low-fidelity evaluation $n$ is set to 200. During the optimization, the population size $N$ is set to 20, and the iteration number $G$ is set to 20. The number of selected individuals that are preserved into the archive $m$ is set to 3. The searched architecture is retrained by standard training on the whole training dataset. The training epoch is set to 50. The trained models are evaluated by 12 types of evaluations, including Clean, Hue, Saturation, Rotation, Brightness, Contrast, CAA, FGSM-$\mathscr{l}_{\infty}$, PGD-$\mathscr{l}_{\infty}$, MI-attack-$\mathscr{l}_{2}$, PGD-$\mathscr{l}_{2}$ and MI-attack-$\mathscr{l}_{2}$, respectively. The detailed parameters of different attacks are listed in Table \ref{attacksetting}. Among that, CAA is the combination of five semantic adversarial attacks and PGD-$\mathscr{l}_{\infty}$, where the magnitude is the same as the single type of attack. Our comparison models include the manually-designed ones and those searched by NAS. The manually-designed models include GoogleNet, MobileNetV2, 
DenseNet121 and VGG19. The models found by NAS are Random \cite{li2020random}, DARTS \cite{liu2018darts}, PDARTS \cite{liu2018progressive}, RACL \cite{dong2020adversarially}, RNAS  \cite{qian2022robust}, and Advrush \cite{AdvRush2021}. All models adopt the same training strategy.

\subsection{Performance of ES-CRNA-ME on CIFAR10}\label{42}

\begin{figure}[htbp]
	\centering
	\includegraphics[width=.53\textwidth]{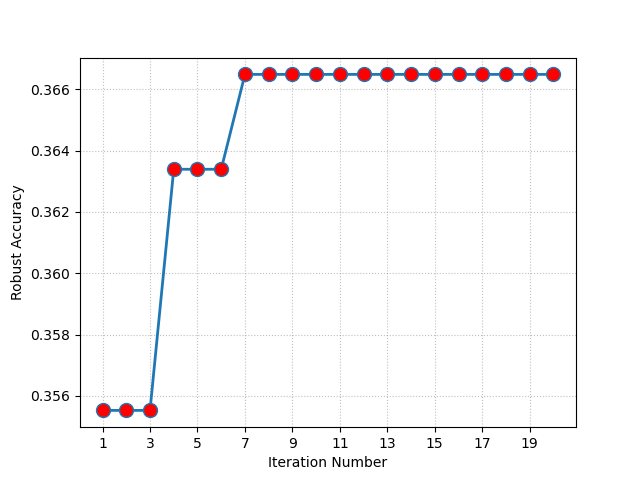}
	\caption{The corelation of the robust accuracies between different types of attacks.}
	\label{fitness}
\end{figure}

\begin{figure}[t]
\centering
\subfigure[Normal Cell]{
\includegraphics[width=.4\textwidth]{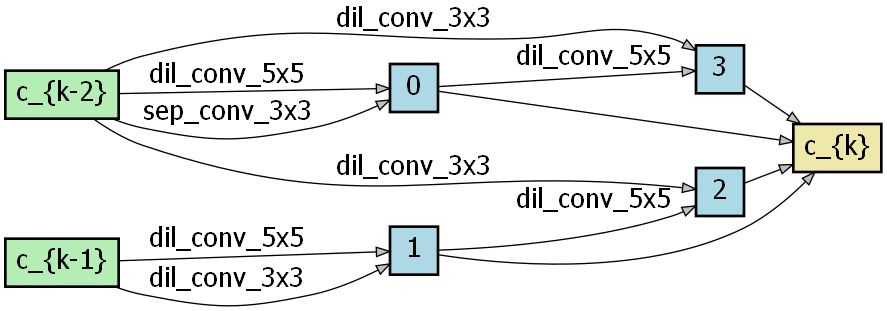}
\label{v10_n}
}
\quad
\subfigure[Reduction Cell]{
\includegraphics[width=.4\textwidth]{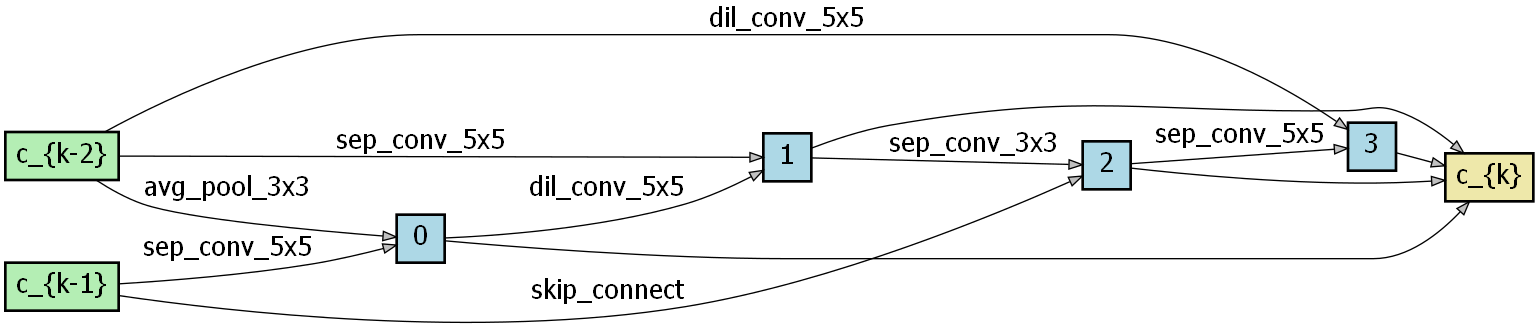}
\label{v10_r}}
\caption{Normal and reduction convolutional cell architectures found by ES-CRNA-ME on CIFAR10 dataset.}
\label{v10}
\end{figure}

We first conduct the experiments on the CIFAR10 dataset. The robust accuracy of the searched architectures on the supernet during the optimization is shown in Figure \ref{fitness}. From Figure \ref{fitness}, we can see that the robust accuracy increase significantly with the increase of the iteration number, which verifies the effectiveness of the whole surrogate model-based optimization. Then we retrain the searched model and evaluate it using multiple types of attacks. The performance of different models is presented in Table \ref{Res10}. From Table \ref{Res10}, we can see that some manually-designed models possess high clean accuracy, while they are not robust enough against diverse adversarial attacks. In the cell-based architectures, the randomly selected one is also not robust. However, by NAS methods, the robustness performance of model architectures can be improved greatly. But achieving the trade-off between the semantic adversarial attacks and $l_{p}$ based attacks is difficult. Taking the comparison between RACL and RNAS as an example, though RNAS is farther more robust on the  $l_{p}$ based attacks, RACL is more robust on semantic attacks. It can prove the necessity of searching for comprehensively robust architectures under multiple types of evaluations. We can also observe that our method can find the most comprehensively robust architectures. The searched architecture is visualized in Figure \ref{v10}.

Besides, the comparison between existing methods and our proposed ES-CRNA-ME is presented in Table \ref{Res100}. From Table \ref{Res100}, we can observe that our method can utilize less computational burden to find better architectures. The comparison methods are differentiable-based architecture search methods, while our proposed method is non-differentiable. In non-differentiable-based search methods, the robustness evaluation is more flexible but generally time-consuming. However, due to the proposed acceleration strategies, our method can efficiently find robust architectures.

\begin{table}[htbp]
	\caption{The comparison of the searching time by different methods.}
	\label{Res100}
	\centering
	\begin{tabular}{cc}
		\hline
		Methods & Time (GPU/Days) \\ \hline
		Advrush       & 1        \\
		DARTS      & 1       \\
		RACL      & 1       \\
		RNAS       & 1  \\
		ES-CRNA-ME            & 0.5     \\ \hline
	\end{tabular}
\end{table}

\subsection{Performance of ES-CRNA-ME on CIFAR100}\label{43}
\begin{figure}[t]
\centering
\subfigure[Normal Cell]{
\includegraphics[width=.5\textwidth]{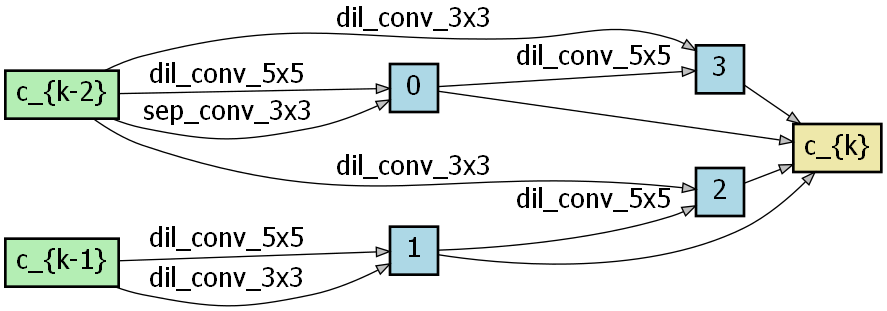}
\label{v100_n}
}
\quad
\subfigure[Reduction Cell]{
\includegraphics[width=.4\textwidth]{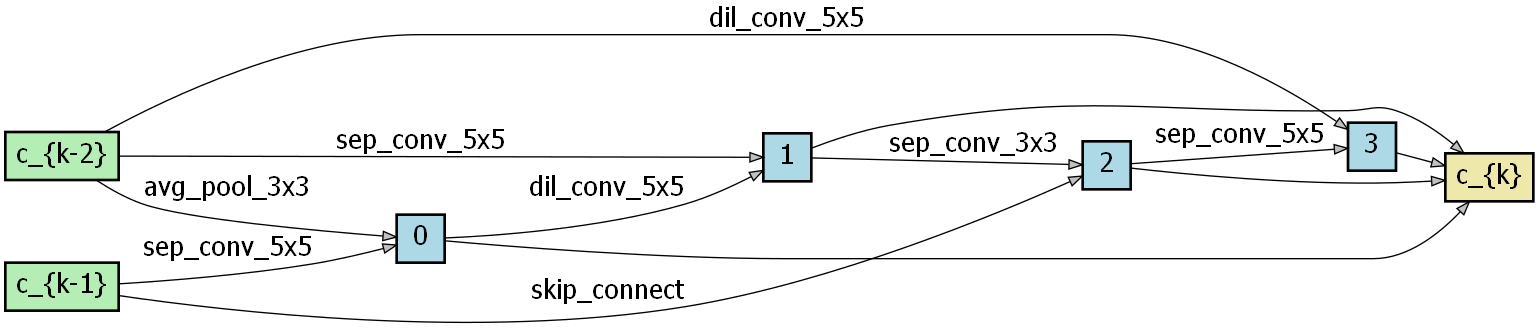}
\label{v100_r}}
\caption{Normal and reduction convolutional cell architectures found by ES-CRNA-ME on CIFAR100 dataset.}
\label{v100}
\end{figure}

We also conduct experiments on the CIFAR100 dataset. The comparison between the performance of the searched model and existing models are listed in Table \ref{results100}. From Table \ref{results100}, it can be seen that our method can still find the mostly comprehensively robust architectures. Compared with manually-designed architectures, the comprehensively robust accuracy of ES-CRNA-ME is farther higher. In addition, ES-CRNA-ME also possesses obvious advantages over randomly selected architectures. Among all comparison methods, PDARTS possess the highest robust accuracy under clean examples and semantic adversarial attacks. However, it has relatively poor performance on $l_{p}$-norm-based attacks. In general, the experimental results on CIFAR100 can also verify the effectiveness of our method. The searched architecture is visualized in Figure \ref{v100}.

\subsection{Analysis of the relativity of different evaluations}\label{44}

\begin{figure}[htbp]
	\centering
	\includegraphics[width=.5\textwidth]{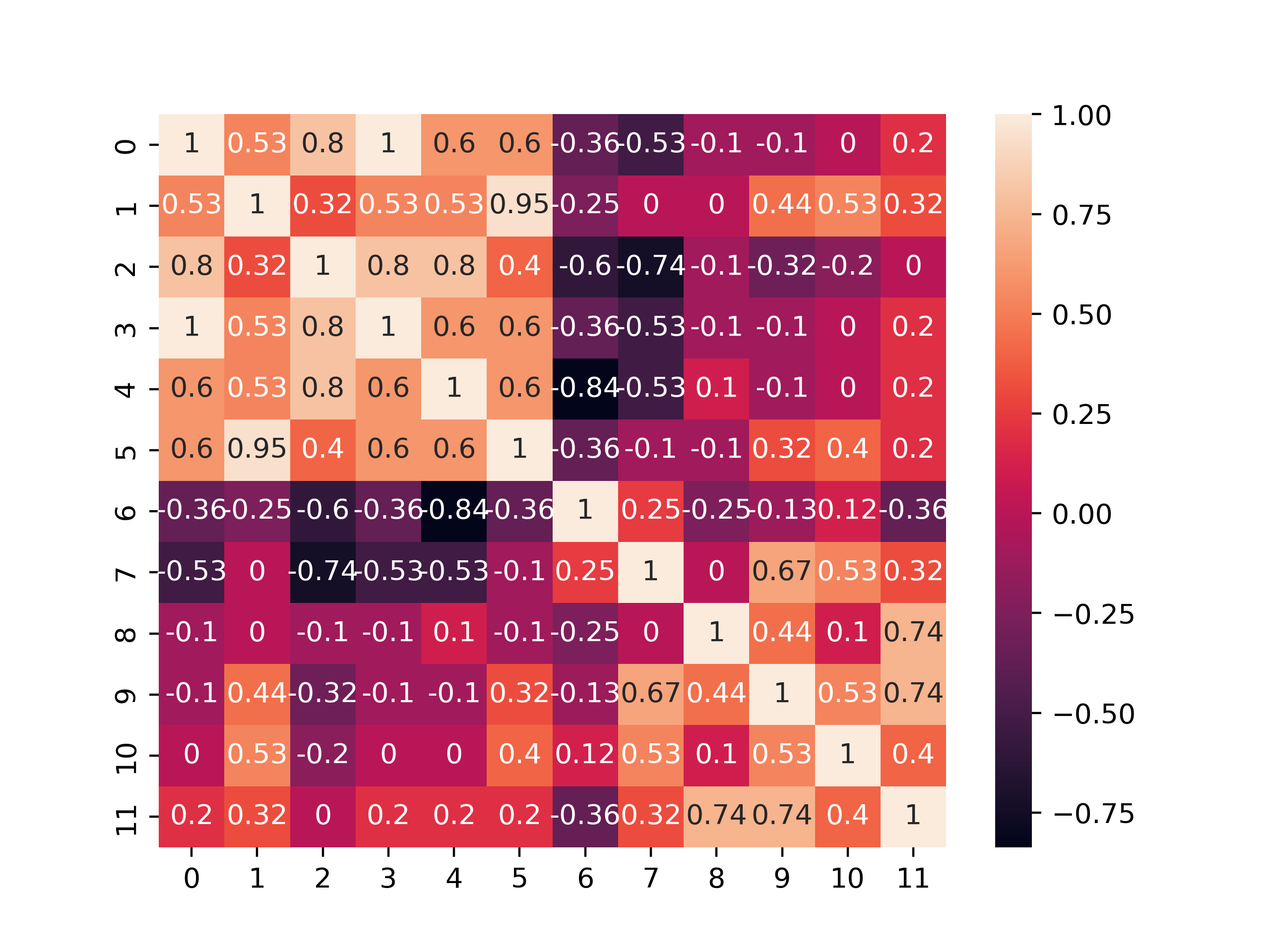}
	\caption{The correlation of the robust accuracies between different types of attacks.}
	\label{flow}
\end{figure}

In this section, the analysis of the relativity of different evaluations on CIFAR10 dataset is performed. After the supernet is trained, we utilize different types of adversarial attacks to evaluate the robust accuracy of the supernet. We sample 5 architectures and calculate the correlation between the accuracies under different evaluations. The result is visualized in Figure \ref{flow}. In this figure, positive values mean positive correlation, while negative values stand for negative correlations. The number from one to twelve represents the robust accuracies under Clean, Hue, Saturation, Rotation, Brightness, Contrast, CAA, FGSM-$\mathscr{l}_{\infty}$, PGD-$\mathscr{l}_{\infty}$, MI attack-$\mathscr{l}_{2}$, PGD-$\mathscr{l}_{2}$ and MI attack-$\mathscr{l}_{2}$, respectively. For instance, the correlation between clean accuracy and Rotation can reach 1. Thus, we just need to evaluate the clean accuracy rather than both of them. We set a threshold, such as 0.7, to select similar evaluations and merge them. The evaluation pairs that meet the constraints are presented in Table \ref{select}. The merged evaluations are shown in Table \ref{mergeresult}. In this way, given the trained supernet and any type of robustness evaluation, we can adaptively merge the similar robustness evaluation and decrease the search cost to efficiently find comprehensively robust neural architectures. In our experiment on the CIFAR10 dataset, we can reduce four robust evaluations. To realize the accurate robustness estimation, three coefficients are needed to be recalculated according to Eq. \ref{coefficient}. We obtain the average coefficient values using five architectures, which are illustrated in Table \ref{select1}. $k_{1}$, $k_{2}$ and $k_{9}$ are the coefficients of three evaluations, including Clean, Hue, PGD-$\mathscr{l}_{\infty}$, respectively. Then, the comprehensively robust accuracy can be obtained according to Eq. \ref{comprehensive} during optimization.

\begin{table}[htbp]
	\caption{The result of merging similar attacks.}
	\label{select}
	\centering
	\begin{tabular}{cc}
		\hline
		Original Attacks & Correlation value \\ \hline
		0-3       & 1       \\
		0-2      &0.8      \\
		1-5&0.95\\
		2-3&0.8\\
		2-4&0.8\\
		8-11&0.74\\
		9-11&0.74\\ \hline
		
	\end{tabular}
\end{table}

\begin{table}[htbp]
	\caption{The result of merging similar attacks.}
	\label{mergeresult}
	\centering
	\begin{tabular}{cc}
		\hline
		Original Attacks & Merged Attacks \\ \hline
		Clean, Saturation, Rotation       & Clean       \\
		Hue, Contrast      & Hue      \\ 
		PGD-$\mathscr{l}_{\infty}$, MI attack-$\mathscr{l}_{2}$ &PGD-$\mathscr{l}_{\infty}$\\ \hline
		
	\end{tabular}
\end{table}

\begin{table}[htbp]
	\caption{The coefficients of merged attacks.}
	\label{select1}
	\centering
	\begin{tabular}{cccc}
		\hline
		Model & $k_{1}$&$k_{2}$& $k_{9}$\\ \hline
		1      & 3.37      &1.84& 1.34\\
		2      &3.37     &1.98&1.71 \\
		3&3.37&2.12&1.52\\
		4&3.51&2.03&1.59\\
		5&3.54&2.10&1.54\\
		Avg &3.43&2.01&1.54\\
		\hline
		
	\end{tabular}
\end{table}

\begin{figure}[htbp]
	\centering
	\includegraphics[width=.48\textwidth]{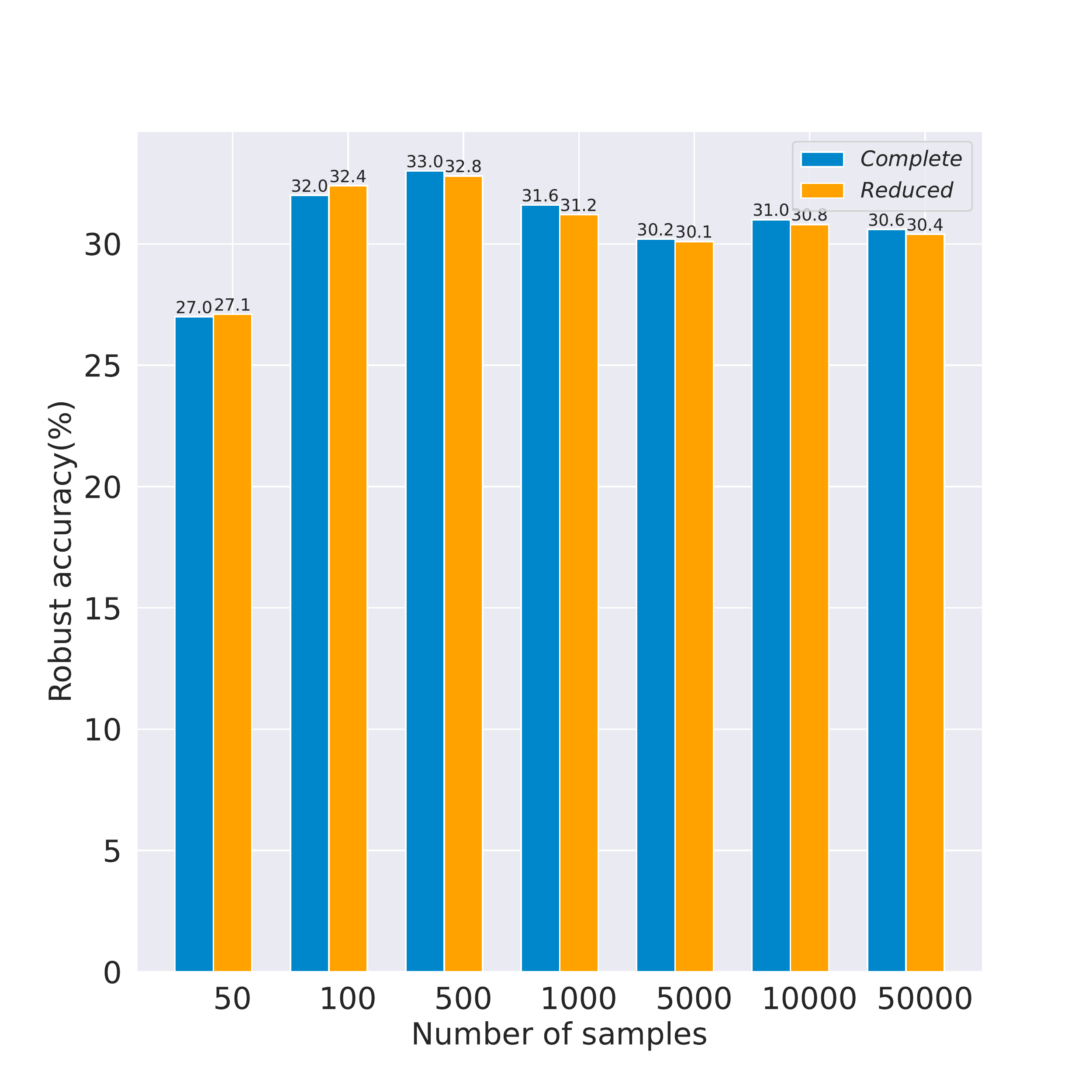}
	\caption{The comparison of robust accuracy using complete and reduced evaluations.}
	\label{robust_acc}
\end{figure}

\begin{figure}[htbp]
	\centering
	\includegraphics[width=.48\textwidth]{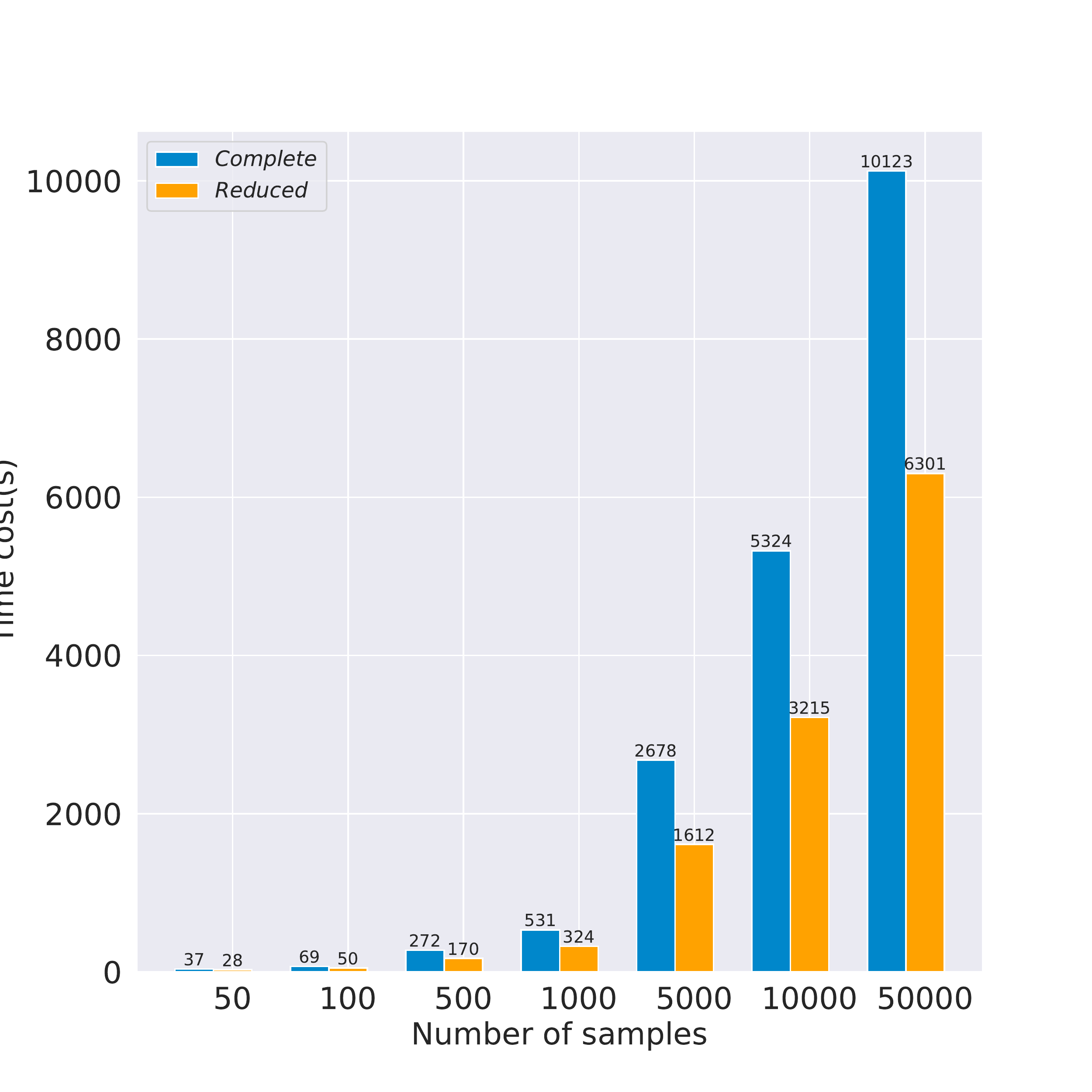}
	\caption{The comparison of time cost using complete and reduced evaluations.}
	\label{time_cost}
\end{figure}

To further illustrate the effectiveness of our proposed reduced evaluations, we select one architecture and evaluate the robust accuracy using complete and reduced robustness evaluation, respectively. The comparison of them is presented in Figure \ref{robust_acc}. From Figure \ref{robust_acc}, we can see that if the number of test samples is too small, the robustness evaluation is not accurate enough. With the increase in the number of test samples, the robustness evaluation tends to be more accurate. In addition, the reduced robustness evaluation can remain at the same level as the complete evaluation. The Kendall’s tau coefficient of them can also reach 1, which verifies the effectiveness of our proposed method. The corresponding evaluation cost is also shown in Figure \ref{time_cost}. From Figure \ref{time_cost}, we can see that our method can reduce over 1/3 time cost to evaluate the model architectures.

\subsection{Analysis of multi-fidelity online learning}\label{45}

\begin{figure}[htbp]
	\centering
	\includegraphics[width=.47\textwidth]{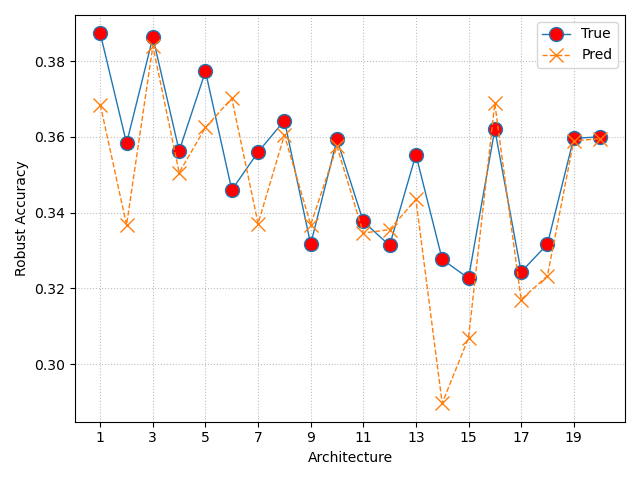}
	\caption{The corelation of the robust accuracies between different types of attacks.}
	\label{surrogate}
\end{figure}

In this section, we present the performance of the surrogate model utilized in our proposed method. First, we evaluate the initial prediction accuracy using low-fidelity data. We sample randomly 20 architectures and evaluate them on the supernet. The comparison of the true robust accuracy and the predicted one is presented in Figure \ref{surrogate}. From Figure \ref{surrogate}, we can see that the surrogate model can possess a superior prediction ability of the performance of different architectures.

By combining the two above-mentioned strategies, we can realize the purpose of an efficient search for comprehensively robust neural architectures. To illustrate the effect of two strategies on the performance of searched architectures, we also further conduct the ablation study. The experimental result is presented in Table \ref{comparison}. S-CRNA utilizes the evolutionary algorithm to find robust architectures under complete evaluations. ES-CRNA uses the reduced evaluation while not utilizing the multi-fidelity surrogate model. In S-CRNA and ES-CRNA, both the population size and iteration number are set to 20. From Table \ref{comparison}, we can see that S-CRNA takes the most computational burden, whose search time can reach 2.5GPU/Days. When the reduced evaluation is performed, that is also, ES-CRNA, the search time can be reduced to 1.5GPU/Days. On the basis of reduced evaluation, when the multi-fidelity online surrogate model is utilized, the whole-time cost can achieve the lowest level, which is only 0.5GPU/Days. In addition, ES-CRNA-ME can possess slightly higher robust accuracy than S-CRNA and ES-CRNA, which can verify the effectiveness of our proposed strategies.

\begin{table}[htbp]
	\caption{The comparison of the searching efficiency of ES-CRNA-ME.}
	\label{comparison}
	\centering
	\begin{tabular}{ccc}
		\hline
		Methods & Time (GPU/Days) & Accuracy\\ \hline
		S-CRNA      & 2.5    & 71.51\\
		ES-CRNA&1.5& 71.38\\
		ES-CRNA-ME            & \textbf{0.5}    & \textbf{71.75}\\ \hline
	\end{tabular}
\end{table}

\subsection{Discussions}\label{46}

In the CIFAR10 and CIFAR100 datasets, our experimental results have verified the effectiveness of the proposed efficient search of comprehensively robust neural architectures via multi-fidelity evaluation. The whole framework is first devised to solve the problem of finding comprehensively robust neural architectures. In our work, two key components, including correlation analysis and multi-fidelity surrogate model online learning, can effectively handle the huge computational burden brought by the comprehensive robustness evaluations. However, in our work, we do not deeply explore and analyze the effect of different search spaces on the performance of NAS. We believe this is one key component to defining a suitable and efficient search space in the areas of searching for robust neural architectures.

\section{Conclusions}\label{sec5}

Different from existing works that only focus on automatically finding robust neural architectures against $l_{p}$ norm-based adversarial attacks, in this paper, we first try to apply the NAS method to search for comprehensively robust neural architectures. To realize an efficient search, we use the weight-sharing neural architecture search to reduce the computational cost of retraining the architecture. In addition, we propose two strategies to reduce the evaluation cost. On the one hand, we use correlation analysis to merge similar attacks, which can save about 1/3 computational burden. On the one hand, the multi-fidelity online surrogate model is utilized to substitute the real evaluation, which can further accelerate the search process. Experimental results on CIFAR10 and CIFAR100 datasets show that our proposed method can find comprehensively more robust neural architectures than existing state-of-art methods. Besides, the time cost of our method is also the least. In future work, the more efficient search space towards improving the robustness against multiple types of attacks can be further studied.

\section*{Acknowledgment}
This work was supported in part by National Natural Science Foundation of China under Grant No.52005505.



\printcredits

\bibliographystyle{model1-num-names}

\bibliography{my_refs}


%
%

\end{document}